# A learning-based solution approach to the application placement problem in mobile edge computing under uncertainty

T.H. Hejazi[1*], Z. Ghadimkhani[1], and A. Borji[1]

[1]*Department of Industrial Engineering, College of Garmsar, Amirkabir University of Technology (Tehran Polytechnic), Tehran, Iran (\* Corresponding author email address: t.h.hejazi@aut.ac.ir)*

## Abstract

Placing applications in mobile edge computing servers presents a complex challenge involving many servers, users, and their requests. Existing algorithms take a long time to solve high-dimensional problems with significant uncertainty scenarios. Therefore, an efficient approach is required to maximize the quality of service while considering all technical constraints. One of these approaches is machine learning, which emulates optimal solutions for application placement in edge servers. Machine learning models are expected to learn how to allocate user requests to servers based on the spatial positions of users and servers. In this study, the problem is formulated as a two-stage stochastic programming. A sufficient amount of training records is generated by varying parameters such as user locations, their request rates, and solving the optimization model. Then, based on the distance features of each user from the available servers and their request rates, machine learning models generate decision variables for the first stage of the stochastic optimization model, which is the user-to-server request allocation, and are employed as independent decision agents that reliably mimic the optimization model. Support Vector Machines (SVM) and Multi-layer Perceptron (MLP) are used in this research to achieve practical decisions from the stochastic optimization models. The performance of each model has shown an execution effectiveness of over 80%. This research aims to provide a more efficient approach for tackling high-dimensional problems and scenarios with uncertainties in mobile edge computing by leveraging machine learning models for optimal decision-making in request allocation to edge servers. These results suggest that machine-learning models can significantly improve solution times compared to conventional approaches.

*Keywords:* Application placement problem; Mobile edge computing; Machine learning; Stochastic programming; Uncertainty.

## Statements and Declarations

Competing interests
The authors declare that they have no known competing financial interests or personal relationships that could have appeared to influence the work reported in this paper.

Funding
This research received no external funding.

Data availability
The data that support the findings of this study are available from the corresponding author, Dr. Taha-Hossein Hejazi, upon reasonable request.

# 1. Introduction

Mobile Edge Computing (MEC) is an innovative technology designed to provide services to users of mobile internet devices. It extends cloud computing services to the edge of the network, specifically to mobile base stations. The MEC platform significantly reduces network delay by enabling computing and storage capacity at the network's edge. Mobile devices can offload computationally intensive tasks like image processing and mobile gaming to these local servers. These servers are strategically placed in parks, bus terminals, shopping malls, and other locations to leverage the benefits of mobile edge computing services, enhancing user satisfaction with their overall experience. However, mobile phone users do not enjoy the same level of satisfaction as desktop device users due to the limited resources of mobile devices, including processing power, battery life, storage capacity, and energy resources (Abbas et al., 2017; Ahmed, Gani, Khan, et al., 2015; Dinh et al., 2013). Mobile gadgets such as phones, tablets, and laptops have evolved(Scheltens et al., 2016) into versatile tools for entertainment, news exploration, social networking, and commercial development. With the recent advancements in mobile cloud computing, many services originally exclusive to desktops are now available on these devices.

Consequently, a substantial amount of data must be transferred between mobile devices and cloud data centers, increasing bandwidth consumption (Ahmed, Gani, Sookhak, et al., 2015). Nevertheless, these devices are subject to limitations such as energy consumption, storage space, and processing power. Vailshery (Vailshery, 2023) predicted that the number of internet-connected devices worldwide will surpass 29.4 billion by 2030. This exponential growth is a challenge for various industries that rely on computationally demanding applications. Solving problems with significant processing power requirements, like stochastic optimization problems (SOP), can be computationally expensive and time-consuming, often impeding project progress.

The following paragraphs introduce some notable MEC research conducted in recent years. Researchers have employed various methodologies to investigate the dynamic nature of MEC systems, primarily driven by user mobility. The rest of the paper is organized as follows. In Section 2, we review the related work. Section 3 defines the problem of placing applications in MEC and the proposed method. In Section 4, we describe the experimental setup and discuss the results. In Section 5, we conclude the paper and suggest possible directions for future works.

# 2. Related works

Optimization has always been a driving force behind the advancement of machine learning theory and a key component of many of its methods (Xu et al., 2016). On the other hand, machine learning models have been used previously as optimization methods as well (Bello et al., 2016; Larsen et al., 2022; Lodi et al., 2020; Mossina et al., 2019; Sun et al., 2022). These models have been employed in various prediction and parameter estimation problems (Abbasi et al., 2008; Londhe & Charhate, 2010). One of the main reasons for using machine learning to solve optimization problems is the immediate need for solutions in practical and large-scale scenarios, where this research has aimed to reduce computation time (Larsen et al., 2022). In this regard, Smith (1999) has investigated the application of neural networks in combinatorial optimization problems. Paliwal and Kumar (2009) have also explored the use of neural networks in various fields, including engineering and construction. Bengio et al. (2021) have reviewed and examined research in operations and combinatorial optimization problems using machine learning. Smith and Gupta (2000) have utilized machine learning algorithms to detect fraud in financial security-related issues. Larsen et al. (2022) used machine learning algorithms to generate optimal or near-optimal solutions for a combinatorial optimization problem in railway scheduling for freight transportation. They demonstrated that deep learning models create relatively accurate predictions in a shorter computational time. Fischetti and Fraccaro (2019) used machine learning models to predict the desirable performance of a wind power plant optimization problem. They needed about 10 hours to find the potential value of a new site for the optimization model. Instead of solving the optimization model, which requires considerable computational time, they employed neural networks and support vector regression models. Bello et al. (2016) employed a neural network approach to achieve solutions close to optimal for a Traveling Salesman Problem (TSP) with a maximum of 100 nodes and a knapsack problem with up to 200 items. They trained a recurrent neural network that predicts distributions over different city permutations based on a set of city coordinates. Mossina et al. (2019) used the SuSPen meta-algorithm as a machine learning tool for predicting a subset of decision variables in a multi-label classification problem. This meta-algorithm aims to expedite the solution of repeated combinatorial problems based on historical data. Lodi et al. (2020) used a binary classification and regression approach to predict whether a new instance of an optimization problem, derived from a reference instance, shares all or part of its optimal solution with the reference instance. They applied their proposed approach to the facility location problem and found that they could achieve good solutions in less time. Václavík et al. (2018) presented an approach to accelerate the computation time of the branch and price algorithm, a precise and powerful method for solving complex combinatorial problems. They show how data from previous pricing problem solutions can be utilized to reduce the solution space of pricing problems in future iterations. Their solution is based on an online machine learning method that uses a very fast regression model. Vinyals et al. (2015) utilized the Pointer Network (Ptr-Net) approach to learn approximate solutions for three geometric problems: convex hull finding, Delaunay triangulation, and the Traveling Salesman Problem (TSP). Abbasi et al. (2020) proposed a method for decision-making regarding the transfer of blood units within a network of hospitals. They compared the learned decisions made by machine learning models with optimal results obtained if the hospitals had access to optimization solutions. The results showed that using a trained model reduced the daily average cost by approximately 29% compared to the current policy. Sun et al. (2022) exclusively focused on the Vehicle Routing Problem (VRP), which combines elements of the Traveling Salesman Problem and the Knapsack Problem. They presented a cooperative approach combining machine learning and ant colony optimization.

Table 1 compares the recent studies using machine learning to solve optimization problems, categorized based on their application, input parameters, and output.

Table 1. A comparison of recent studies on the use of machine learning to solve optimization problems

| Work | Year | Problem | Input | Output | Learning algorithm |
|---|---|---|---|---|---|
| (Vinyals et al., 2015) | 2015 | TSP and the Dylan triangle | Coordinates of points | A sequence of points | ANN |
| (Bello et al., 2016) | 2016 | TSP and knapsack problem | Coordinates of points for traveling salesman and value and weight of items for knapsack | A sequence of points for the traveling salesman and items for the knapsack | ANN |
| (Václavík et al., 2018) | 2018 | Time-sharing and scheduling of nurses' problem | Sample characteristics (number of days, shifts, skills, etc.) | Scheduling of nurses | FS and regression |
| (Larsen et al., 2022) | 2018 | Load planning problem | parameters of definite operational problems (such as the number of rail cars of each type and the number of containers, etc.) | Value of decision variables using aggregation and sub-selection methods | Regression and ANN |
| (Mossina et al., 2019) | 2019 | Multi-label classification problem | Problem parameters (such as demand or distance matrix and calculated relative optimal path) | Solutions for a subset of decision variables | Suspension algorithm |
| (Lodi et al., 2020) | 2020 | Facility location problem (diagnosing the malfunction in the airport system of a region and providing a solution) | Problem parameters (possibilities available in reference solutions) | Need to reroute or reschedule | decision tree, ANN, LR, and simple Bayesian |

| (Abbasi et al., 2020) | 2020 | Blood supply chain | Blood inventory levels | Decision variables related to orders and shipping | ANN, KNN, decision tree, and RF |
|---|---|---|---|---|---|
| (Sun et al., 2022) | 2020 | Orientation problem | Specific characteristics of the problem and statistical criteria | Edge optimization | Neural graph networks, logistic regression, and SVMs |
| Proposed research | 2023 | Allocation and placement problem | The distances of each user from each server and the number of user requests | Server assignment | MLP and SVMs |

*Note:* ANN, artificial neural network; TSP, Traveling Salesman Problem; FS, Feature selection; LR, logistic regression; KNN, K nearest neighbor; RF, random forest; SVM, support vector machines; MLP, Multi-layer perceptron.

### 3. Methodology

This study employs machine-learning models to predict the optimal application placement in edge computing, intending to comprehend the correlation between user and server locations to allocate user requests effectively. By leveraging machine-learning techniques, these models anticipate the allocation of user requests by considering the distance between users and all available servers. Two machine-learning models have been trained using input parameters to address the stochastic optimization model in an uncertain position. While the hyperparameters are typically selected empirically, the models learn the parameters of the machine-learning models during the training process. In this study, the Grid search algorithm is utilized to search and determine the ideal hyperparameter values.

Additionally, the cross-validation algorithm is employed to modify the critical hyperparameters of the machine-learning models. Considering all the provided hyperparameters for the search, a network search is conducted based on a full factorial design of experiments (Hsu et al., 2003). Working with machine-learning models poses several challenges, primarily due to the inherent uncertainty surrounding their performance until they are rigorously tested on independent datasets. To address this, we employ K-fold validation, a widely recognized cross-validation technique, to augment our machine-learning model's efficacy. In our research, we leverage the power of 5-fold validation, a robust approach that enables us to attain optimal outcomes and significantly advance our findings.

#### 3.1. Description of the studied system

The MEC system comprises a set of edge servers that offer user services. In this setup, the edge servers are located within a base station, collectively providing computing capacity equivalent to the maximum number of unit-sized containers that the server can accommodate. Users interact with the system through a user interface and can access it via Wi-Fi, 4G, or 5G networks. They can make requests to deploy small applications on the edge servers. For this research, all servers are assumed to be active and accessible. Each application is designed to cater to a single user request. The users' locations are determined by their two-dimensional network coordinates. While users can change their locations between time slots, they cannot do so within a time slot. As shown in Figure 1, during the second stage, users can move to nine different modes, either moving to adjacent cells or remaining in the same cell. The second stage of the optimization model employs each user's stochastic scenarios to explore various possibilities for their subsequent actions. All servers are considered operational and reachable. The users' two-dimensional grid coordinates are a reference for their current positions, which remain fixed. However, users can change their locations between time slots and move to any neighboring cell.

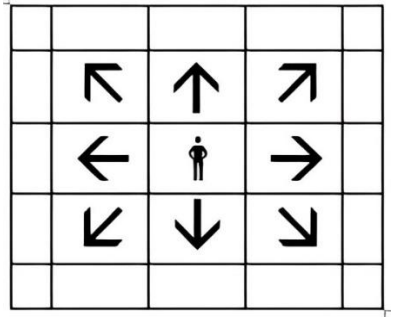

Fig1. User movement scenarios

The primary objective of this problem is to maximize the overall Quality of Service (QoS) of the system while considering the limitations on energy consumption and processing power of the edge servers. A two-stage stochastic integer programming is developed to address this challenge. Notations utilized in the problem formulation are illustrated in Table 2.

Table 2. Notation

| Notation | Description |
|---|---|
| U | A set of users |
| S | A set of servers |
| $\xi$ | A set of scenarios about user movement $\{SC_1, \ldots, SC_E\}$ |
| $QOS^1_{ij}$ | The quality of service that the user $U_i$ receives from the server $S_j$ in the first step. |
| $QOS^2_{ij(\xi)}$ | The quality of service that the user $U_i$ receives from the server $S_j$ in the second step is based on each scenario. |
| $E_j$ | Server $S_j$ energy budget |
| $\epsilon_{ij}$ | The amount of energy consumed by the request of the user $U_i$ when running on the server $S_j$ |
| $\sigma$ | The fixed coefficient in the calculation of $\epsilon_{ij}$ |
| $Q_j$ | The computing capacity of the $S_j$ server |
| $R_i$ | The size of the container requested by the user $U_i$ |
| $d_{ij}$ | Distance between user $U_i$ and server $S_j$ in the first step. |
| $d^2_{ij(\xi)}$ | The distance between the user $U_i$ and the server $S_j$ in the second stage is based on each scenario. |
| $\rho_{j'j}$ | Cost of moving associated with changing user request allocation from server $S_{j'}$ to server $S_j$ |
| $\gamma$ | The fixed coefficient in the calculation of $QOS_{ij}$ |
| $\mathbb{E}_\xi[.]$ | The expected value of the recourse function for user navigation scenarios. |
| $x1_{ij}$ | The binary variable of the first stage is 1 if the request of the user $U_i$ is assigned to the server $S_j$; otherwise, it is 0. (First stage decision variable) |
| $x^2_{ij(\xi)}$ | The binary variable of the second step is 1 if the application of the user $U_i$ is assigned to the server $S_j$ in each scenario; otherwise, it is 0. (Second stage decision variable) |
| $y2_{ij'j(\xi)}$ | The binary variable of the second stage equals 1 if the user application $U_i$ is transferred from server $S_{j'}$ to server $S_j$ in the second stage; otherwise, it is zero. |

In this system, the QoS is influenced by two factors: the distance between the user and the server and the magnitude of the user's request. The distance between users and servers is assumed to be calculated using the Manhattan distance method. Consequently, when a user's request is assigned to an edge server in closer proximity, the user experiences reduced latency and improved QoS. The relationship defined as equation 1 represents the distance between the user and the server.

$$QOS_{ij} = \frac{\gamma . R_i}{d_{ij}} \qquad (1)$$

The energy budget of the server $S_j$ is denoted by $E_j$. Equation 2 shows the amount of energy consumed for the request of the user $U_j$ on the server $S_j$; moreover, $\sigma$ is a constant coefficient in this equation.

$$\epsilon_{ij} = \frac{\sigma \cdot R_i}{Q_j} \qquad (2)$$

### 3.1.1. Mathematical representation of the problem of placement of applications in MEC

The following is the formulation of the MEC two-stage stochastic programming placement problem, which draws inspiration from the model [22] but incorporates certain modifications:

$$\max_{X,Y} \sum_{i \in U} \sum_{j \in S} \frac{\gamma \cdot R_i}{d^{\gamma}_{ij}} \cdot x^1_{ij} + \mathbb{E}_{\xi}\left[\sum_{i \in U} \sum_{j \in S} \left(\frac{\gamma \cdot R_i}{d^{\gamma}_{ij(\xi)}} \cdot x^2_{ij(\xi)} - \sum_{j' \in S} \rho_{j'j} \cdot y^2_{ij'j(\xi)}\right)\right] \qquad (3)$$

The objective function (3) comprises three distinct components. The first component aims to maximize the overall quality of service received during the initial phase. The second component focuses on optimizing service quality during the second stage. Lastly, the third component accounts for changing the server in the second stage. Essentially, the objective function aims to maximize the quality of the entire service in both stages while considering the deduction of costs incurred from server changes in the second stage.

Subject to:

$$\sum_{i \in U} \frac{\sigma \cdot R_i}{Q_j} \cdot x^1_{ij} \leq E_j \qquad \forall j \in S \qquad (4)$$

$$\sum_{i \in U} \frac{\sigma \cdot R_i}{Q_j} \cdot x^2_{ij(\xi)} \leq E_j \qquad \forall j \in S, \xi \in SC \qquad (5)$$

Constraints (4) and (5) prevent the server from exceeding its energy budget when overloaded.

$$\sum_{j \in S} x^1_{ij} \leq 1 \qquad \forall i \in U \qquad (6)$$

$$\sum_{j \in S} x^2_{ij(\xi)} \leq 1 \qquad \forall i \in U, \xi \in SC \qquad (7)$$

Constraints (6) and (7) ensure that user requests are assigned to no more than one server per step.

$$\sum_{j \in S} x^1_{ij} \leq \sum_{j \in S} x^2_{ij(\xi)} \qquad \forall i \in U, \xi \in SC \qquad (8)$$

Constraint (8) provides the system's uninterrupted processing of user requests.

$$x^2_{ij(\xi)} + x^1_{ij'} - 1 \leq y2_{ij'j(\xi)}$$

$$\forall i \in U, j \in S, j' \in S, \xi \in SC ; j \neq j' \qquad (9)$$

Constraint (9) provides that the decision variables are adjusted accordingly when the user server is moved and changed.

$$x^1_{ij}, x^2_{ij(\xi)}, y2_{ij'j(\xi)} \in \{0,1\} \qquad \forall i \in U, j \in S, j' \in S, \xi \in SC \qquad (10)$$

As shown in Equation (10), decision variables are binary.

#### 3.1.1.1. Solving the problem of placing applications in MEC

Integer programming is widely regarded as one of the most popular modeling techniques for tackling combinatorial optimization problems. In numerous applications, the solution of an integer programming model involves preserving the model's inherent structure and appearance while iteratively adjusting and generating random parameters. These recorded solutions subsequently offer a valuable opportunity for machine learning to uncover the intricate relationships between model structures and solution values.

To visualize this process, Figure 2 showcases the input and output of a machine-learning model employed to predict the solutions of an optimization model in the context of mobile edge computing.

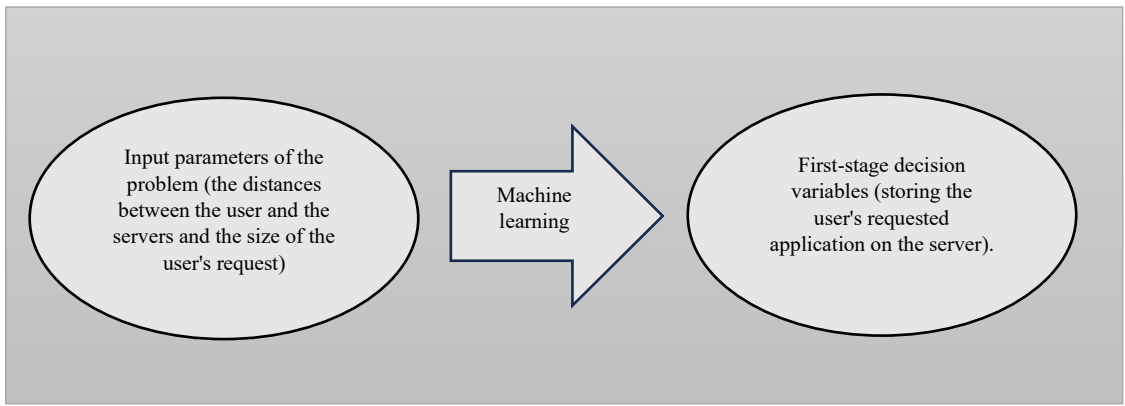

Fig 2. Predicting the solution of an application placement optimization problem in MEC through machine learning

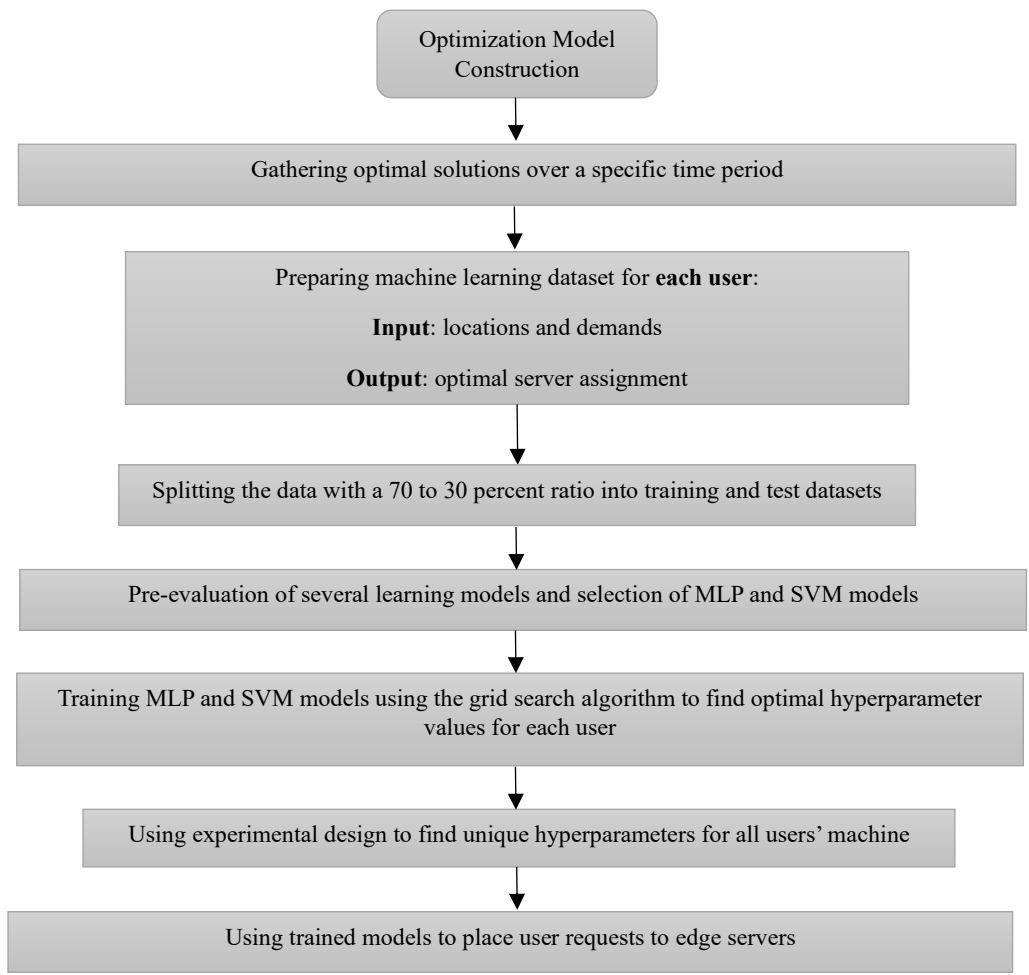

Fig. 3. The overall procedure of the proposed method

### 3.2. Machine learning phase

This section examines the application of neural networks and their role in classification. Furthermore, it provides a detailed description of the machine-learning methods employed in the study. Various machine-learning models have been used, and following a pre-evaluation, the SVM and MLP algorithms have been selected for a comprehensive analysis, which is discussed in the subsequent sections. Figure 3 illustrates the complete workflow of the proposed methodology.

### 3.2.1. Multi-layer perceptron

The Multi-layer perceptron (MLP) is a neural model that aims to emulate the intricate workings of the human brain. It achieves this by closely examining the network behavior and signal propagation observed in the human brain. Like in the brain, each neuron within an MLP receives an input, processes it, and transmits the processed information to other neurons. This sequential and interconnected behavior ultimately leads to various outcomes such as decision-making, actions, and different desired results. To help visualize the concept, Figure 3 illustrates a simplified MLP network.

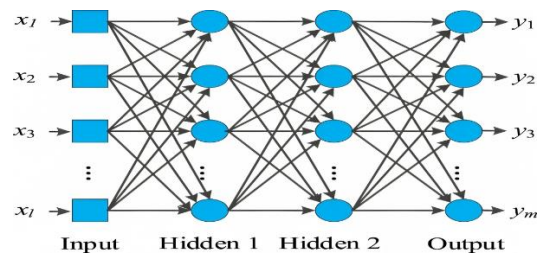

Fig 4. The architecture of a Multi-layer perceptron [28]

Figure 4 illustrates a Multi-layer perceptron (MLP), a type of neural network connecting multiple layers in a directed graph. In this architecture, the signal flows in a unidirectional manner through the nodes. While the input nodes transmit the data, each subsequent node applies a nonlinear activation function except for the input layer. The MLP consists of an input, hidden, and output layer. Multiple hidden layers make the MLP a form of deep learning. It enables the network to learn complex, nonlinear relationships in the data.

In contrast, a single-layer perceptron is more suitable for basic and linear classifications. Therefore, the choice between single-layer and Multi-layer perceptron depends on the nature of the classification task. The perceptron algorithm, used in MLPs, is a supervised learning algorithm that requires labeled records for training. It operates as a linear algorithm, performing sequential and individual identification operations. Each neuron in the perceptron neural network is assigned specific weights for its inputs. Through an automatic learning process, the network adjusts these weight coefficients to minimize the error rate. This adjustment is achieved by comparing the calculated network output with the given label and updating the weights iteratively until convergence. In Equation 11, typically, the input is represented by a feature vector x, which is multiplied by the weight vector w, and the bias term b is added. This equation represents the computation performed at each neuron in the MLP.

$$y = w * x + b \qquad (11)$$

A perceptron generates a single output from multiple inputs with real-valued weights by constructing a linear combination. In Equation (12), it is mathematically stated:

$$y = \varphi(\sum_{i=1}^{n} w_i x_i + b) = \varphi \qquad (12)$$

Where w is the vector of weights, x is the vector of inputs, b is the bias, and φ is the nonlinear activation function.

### 3.2.1.1. MLP learning method steps

Step 1: The data starts at the input layer and propagates through the neural network to reach the output layer. This process is known as forward propagation.

Step 2: Once the output is obtained, the error is calculated by comparing the predicted and actual results. The goal is to minimize this error.

Step 3: Backpropagation is performed to adjust the neural network weights. It involves finding the derivative of the error with respect to each weight in the network and using it to update the model.

These three steps are repeated multiple times for the model to learn the optimal weights. Finally, the output passes through a threshold function to determine the predicted class labels.

In an MLP neural network, the number of layers, neurons, and alpha parameters are referred to as hyperparameters. These hyperparameters need to be adjusted, as described in the following paragraphs. Alpha can be used to control the likelihood of "overfitting" and "underfitting." Increasing alpha can help mitigate high variance (an indication of overfitting) by promoting smaller weights, which results in a less complex decision boundary.

### 3.2.2. Support Vector Machines

Support Vector Machines (SVMs) are powerful algorithms that can effectively handle complex datasets by representing them on a single plane and allowing for multiple boundaries between labels (Borji et al., 2023). In cases where a straight line cannot classify the records accurately, SVMs transfer the problem to a higher-dimensional space, such as 3D (Wang et al., 2010). When it comes to multi-class problems, SVMs employ the one-to-one technique. This approach utilizes a series of two-class classifiers, each responsible for distinguishing between two classes. For instance, one classifier determines if a new sample belongs to the first or second class, while another determines if it belongs to the first or third class.

In contrast, alternative classifiers solve two-class classification by utilizing voting to assign the new sample to the class with the most votes. Due to the presence of six classes, this study requires fifteen two-class classifications using the one-to-one technique. In Perceptron neural networks, the primary goal is to enhance the network structure to minimize model error. However, in the case of support vector machines, the primary objective is to mitigate operational risk associated with model failures. As previously mentioned, this research aims to expedite the solution to the problem of placing applications and assigning user requests to mobile edge servers. This is achieved by utilizing machine learning techniques that address the inherent uncertainty of the problem.

## 4. Results and comparative study

After conducting initial tests on well-known machine learning models, MLP and SVM models have been utilized to develop an optimization model. These machine learning models play a crucial role in making decisions regarding allocating user requests to servers, particularly in addressing the challenge of application placement on edge servers. However, it is imperative to initially determine the association between users and servers to facilitate future computation of transfer charges in the event of server relocation.

### 4.1. The structure of the problem of placing applications in edge servers

The initial locations of 20 users and five homogeneous servers are randomly determined in a 20×20 two-dimensional grid using the uniform function in GAMS software. Depending on the requested applications, the amount of user requests varies between 1 and 10 containers and is determined randomly (Badri et al., 2019). The Manhattan formula is employed to measure the distance between users and servers and assess the level of service users receive from servers over a two-time period, calculated using Equation (1).

In the second stage, users can move to nearby cells or remain in their current cell. The problem was approached using a different number of scenarios (|SC|), ranging from 5 to 50, to analyze changes in the objective function. After considering these scenarios, it was determined that 25 would be chosen due to the marginal difference of only 0.2% in the objective functions at |SC|=25. To address this, 25 movement possibilities are generated randomly for each user. Server allocation differs in various circumstances, such as holidays, conferences, or events that lead to people gathering in specific geographic areas, compared to modes where people are scattered among the servers. Therefore, the placement of the applications requested by users has been examined in three cases.

1. The default situation represents a uniform distribution of users across the entire network, with a 20×20 grid. (Figure 5 illustrates the 1800 training records.)
2. In a special scenario, user dispersion may be concentrated in a specific area. (Figure 6 displays the 1800 training records.)
3. Another situation involves a mixture of the distributions above, where users are dispersed throughout the entire network while also clustering in a specific area with a high concentration. (Figure 7 showcases the 3600 training samples.)

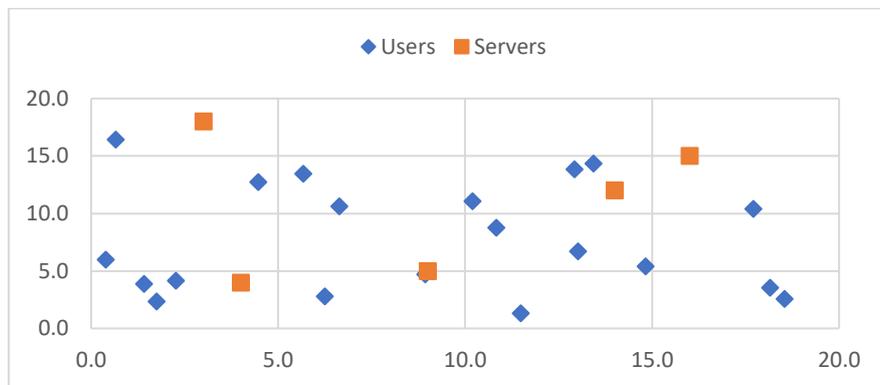

Fig 5. A hypothetical representation of the location of users and servers in the normal mode

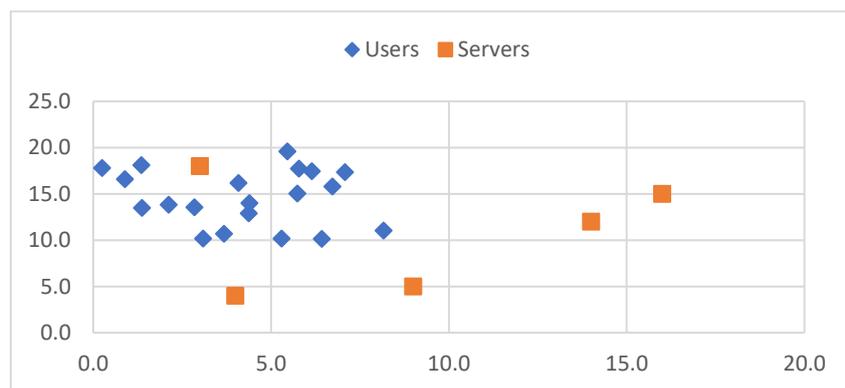

Fig 6. A hypothetical representation of the location of users and servers in a special mode

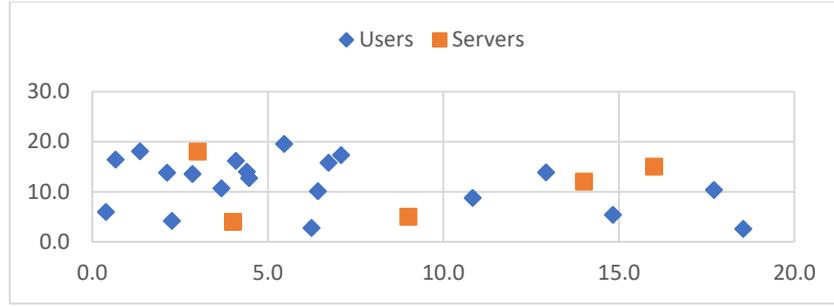

Fig 7. Hypothetical location of users and servers in the mixed mode

It should be noted that some fixed parameters in this paper are shown in Table 3.

Table 3. Fixed parameters of the problem

| Parameter | $\sigma$ | $\gamma$ | $E_j$ | $Q_j$ |
|---|---|---|---|---|
| Value | 396 | 100 | 396 | 24 |

In this research, all the methods used have been carried out in the same configuration has been adopted to address the optimization problem and determine the fixed parameters essential for Equations (1), (2), (4), and (5).

As mentioned before, it is a realistic assumption that historical data about the running optimization models are available for training the learning machines. However, in this study, we generate hypothetical datasets. Machine learning models are then employed to predict the first stage's decision variables, which involves allocating users to servers. A sufficiently large number (N=1800) of instances are generated and solved. Input features such as the distances of users from existing servers and the volume of their requests are utilized to determine the optimal placement of applications. After solving the model by testing various input parameters, the test and training datasets are created from the processed dataset. This iterative process involves splitting the modeling dataset into two parts:

1. The training dataset is used to estimate the model's parameters and build the model.
2. The testing dataset is used to evaluate the model's effectiveness.

### 4.2. Choice of hyperparameters

After numerous iterations, the optimal number and sizes of hidden layers for the MLP model were determined. Subsequently, the SVM and MLP model results have been compared in the normal mode, where grid search is employed to identify the optimal hyperparameter values for each user. Table 4 presents the outcomes obtained from this analysis for the test subset.

Table 4. Test accuracy of SVM and MLP for the normal situation using grid search to find each user's optimal hyperparameters.

| Model | SVM | | | MLP | | |
|---|---|---|---|---|---|---|
|  | Accuracy (%) | kernel | C | Accuracy (%) | Hidden_layer_sizes | Alpha |
| User 1 | 92 | LINEAR | 100 | 92 | (256, 128, 64, 32, 16, 8, 6) | 0.001 |
| User 2 | 90 | LINEAR | 100 | 89 | (256, 128, 64, 32, 16, 8, 6) | 0.001 |
| User 3 | 89 | LINEAR | 100 | 90 | (256, 128, 64, 32, 16, 8, 6) | 0.001 |
| User 4 | 90 | LINEAR | 100 | 89 | (256, 128, 64, 32, 16, 8, 6) | 0.001 |
| User 5 | 89 | LINEAR | 100 | 88 | (256, 128, 64, 32, 16, 8, 6) | 0.00001 |
| User 6 | 90 | LINEAR | 100 | 88 | (256, 128, 64, 32, 16, 8, 6) | 0.001 |

| User 7 | 85 | LINEAR | 100 | 88 | (256, 128, 64, 32, 16, 8, 6) | 0.001 |
| User 8 | 90 | LINEAR | 100 | 86 | (128, 64, 32, 16, 8, 4) | 0.001 |
| User 9 | 91 | LINEAR | 100 | 90 | (256, 128, 64, 32, 16, 8, 6) | 0.001 |
| User 10 | 88 | LINEAR | 100 | 89 | (256, 128, 64, 32, 16, 8, 6) | 0.001 |
| User 11 | 91 | LINEAR | 100 | 91 | (256, 128, 64, 32, 16, 8, 6) | 0.001 |
| User 12 | 91 | LINEAR | 100 | 88 | (256, 128, 64, 32, 16, 8, 6) | 0.001 |
| User 13 | 89 | LINEAR | 100 | 90 | (256, 128, 64, 32, 16, 8, 6) | 0.001 |
| User 14 | 92 | LINEAR | 100 | 93 | (256, 128, 64, 32, 16, 8, 6) | 0.00001 |
| User 15 | 88 | LINEAR | 100 | 87 | (256, 128, 64, 32, 16, 8, 6) | 0.001 |
| User 16 | 91 | LINEAR | 100 | 76 | (128, 64, 32, 16, 8, 4) | 0.00001 |
| User 17 | 90 | LINEAR | 100 | 91 | (256, 128, 64, 32, 16, 8, 6) | 0.001 |
| User 18 | 90 | LINEAR | 100 | 90 | (256, 128, 64, 32, 16, 8, 6) | 0.00001 |
| User 19 | 87 | LINEAR | 100 | 88 | (256, 128, 64, 32, 16, 8, 6) | 0.00001 |
| User 20 | 88 | LINEAR | 100 | 89 | (256, 128, 64, 32, 16, 8, 6) | 0.00001 |

Based on the results obtained from the MLP and SVM models, the average accuracy in the normal situation is 89%. In the best-case scenario, the MLP model achieved an accuracy of 93%, while in the worst-case scenario, the accuracy dropped to 76%. Among the users, the most frequently observed accuracy rate is 88%, which occurred five times. Moving on to the SVM model, when examining the results for the normal mode, it can be observed that users achieved an average accuracy of 90%, which is repeated six times on average. The best-case accuracy for the SVM model was 92%, while the worst-case accuracy was 85%.

In the second case, the grid search algorithm was employed to determine the optimal hyperparameters for a special situation. The results obtained for each user are as follows:

Table 5. Test accuracy SVM and MLP for the special situation using grid search to find each user's optimal hyperparameters.

| Model | SVM | | | MLP | | |
|---|---|---|---|---|---|---|
|  | Accuracy (%) | Kernel | C | Accuracy (%) | Hidden_layer_sizes | Alpha |
| User 1 | 84 | LINEAR | 100 | 84 | (256, 128, 64, 32, 16, 8, 6) | 0.001 |
| User 2 | 86 | LINEAR | 100 | 82 | (128, 64, 32, 16, 8, 4) | 0.001 |
| User 3 | 83 | LINEAR | 100 | 85 | (128, 64, 32, 16, 8, 4) | 0.001 |
| User 4 | 83 | LINEAR | 100 | 81 | (128, 64, 32, 16, 8, 4) | 0.00001 |
| User 5 | 80 | LINEAR | 100 | 82 | (256, 128, 64, 32, 16, 8, 6) | 0.001 |
| User 6 | 84 | POLY | 100 | 81 | (128, 64, 32, 16, 8, 4) | 0.00001 |
| User 7 | 85 | LINEAR | 100 | 85 | (256, 128, 64, 32, 16, 8, 6) | 0.001 |
| User 8 | 82 | LINEAR | 100 | 84 | (256, 128, 64, 32, 16, 8, 6) | 0.001 |
| User 9 | 85 | LINEAR | 100 | 86 | (256, 128, 64, 32, 16, 8, 6) | 0.001 |
| User 10 | 82 | LINEAR | 100 | 85 | (128, 64, 32, 16, 8, 4) | 0.001 |
| User 11 | 85 | LINEAR | 100 | 82 | (256, 128, 64, 32, 16, 8, 6) | 0.00001 |
| User 12 | 86 | LINEAR | 100 | 85 | (256, 128, 64, 32, 16, 8, 6) | 0.001 |
| User 13 | 85 | POLY | 100 | 81 | (256, 128, 64, 32, 16, 8, 6) | 0.001 |
| User 14 | 85 | LINEAR | 100 | 83 | (128, 64, 32, 16, 8, 4) | 0.001 |
| User 15 | 86 | LINEAR | 100 | 86 | (256, 128, 64, 32, 16, 8, 6) | 0.00001 |
| User 16 | 84 | LINEAR | 100 | 83 | (256, 128, 64, 32, 16, 8, 6) | 0.001 |
| User 17 | 82 | LINEAR | 100 | 82 | (256, 128, 64, 32, 16, 8, 6) | 0.00001 |
| User 18 | 87 | LINEAR | 100 | 84 | (256, 128, 64, 32, 16, 8, 6) | 0.00001 |
| User 19 | 83 | LINEAR | 100 | 83 | (256, 128, 64, 32, 16, 8, 6) | 0.001 |
| User 20 | 83 | LINEAR | 100 | 81 | (256, 128, 64, 32, 16, 8, 6) | 0.001 |

The performance of SVM and MLP models have been evaluated in a special case using grid search to determine the optimal hyperparameters for each user. The MLP model achieves an average accuracy rate of 82%, with the best case reaching 86% and the worst case at 81%, based on four repetitions across users. On the other hand, the SVM model yields accuracy rates ranging from 84% in the best case to 87% in the worst case, with the lowest accuracy of 80% and the highest repetition rate of 85%. The grid search algorithm produced the following results for the third mode, which combines the first two modes.

Table 6. Test accuracy SVM and MLP for the mixed situation for each user from grid search to find optimal hyperparameters.

| Model | SVM | | | MLP | | |
|---|---|---|---|---|---|---|
| | Accuracy (%) | Kernel | C | Accuracy (%) | Hidden_layer_sizes | Alpha |
| User 1 | 83 | LINEAR | 800 | 83 | (256, 128, 64, 32, 16, 8, 6) | 0.00001 |
| User 2 | 85 | POLY | 100 | 85 | (256, 128, 64, 32, 16, 8, 6) | 0.001 |
| User 3 | 83 | POLY | 100 | 86 | (256, 128, 64, 32, 16, 8, 6) | 0.00001 |
| User 4 | 84 | LINEAR | 1000 | 87 | (256, 128, 64, 32, 16, 8, 6) | 0.001 |
| User 5 | 84 | POLY | 100 | 84 | (256, 128, 64, 32, 16, 8, 6) | 0.001 |
| User 6 | 82 | LINEAR | 400 | 85 | (256, 128, 64, 32, 16, 8, 6) | 0.001 |
| User 7 | 80 | LINEAR | 400 | 82 | (256, 128, 64, 32, 16, 8, 6) | 0.001 |
| User 8 | 80 | LINEAR | 800 | 85 | (256, 128, 64, 32, 16, 8, 6) | 0.001 |
| User 9 | 83 | POLY | 100 | 15 | (256, 128, 64, 32, 16, 8, 6) | 0.00001 |
| User 10 | 80 | LINEAR | 400 | 84 | (128, 64, 32, 16, 8, 4) | 0.001 |
| User 11 | 82 | POLY | 100 | 86 | (256, 128, 64, 32, 16, 8, 6) | 0.00001 |
| User 12 | 83 | POLY | 100 | 85 | (256, 128, 64, 32, 16, 8, 6) | 0.001 |
| User 13 | 85 | POLY | 100 | 86 | (256, 128, 64, 32, 16, 8, 6) | 0.00001 |
| User 14 | 82 | LINEAR | 400 | 83 | (256, 128, 64, 32, 16, 8, 6) | 0.001 |
| User 15 | 84 | LINEAR | 800 | 84 | (256, 128, 64, 32, 16, 8, 6) | 0.001 |
| User 16 | 81 | LINEAR | 1000 | 85 | (256, 128, 64, 32, 16, 8, 6) | 0.00001 |
| User 17 | 84 | POLY | 100 | 86 | (256, 128, 64, 32, 16, 8, 6) | 0.00001 |
| User 18 | 85 | POLY | 100 | 85 | (256, 128, 64, 32, 16, 8, 6) | 0.001 |
| User 19 | 82 | POLY | 100 | 84 | (256, 128, 64, 32, 16, 8, 6) | 0.00001 |
| User 20 | 79 | LINEAR | 100 | 85 | (256, 128, 64, 32, 16, 8, 6) | 0.00001 |

The accuracy rate of 85% has been repeatedly mentioned by users, with eight instances, when discussing the results obtained in the MLP model for the mixed mode. In the best-case scenario, the MLP model achieved an average accuracy of 85%, with the highest accuracy recorded at 87% and the lowest at 82%. On the other hand, the SVM model yields an average accuracy of 83%. The SVM model's highest accuracy is 85%, while the lowest accuracy is 79%. Notably, the accuracy of 83% receives the highest number of user repetitions, with four instances.

Analyzing the results across all three modes, it becomes evident that the MLP model performs better in the mixed mode, while the SVM model performs better in the remaining two modes. This observation is significant considering that the grid search algorithm is employed in this method to find hyperparameters for each user individually, aiming to achieve optimal accuracy levels. These results showed no clear superiority between the accuracy of MLP and SVM. On the other hand, we know that SVM is much simpler in computation and interpretation. So, we choose SVM as our final algorithm.

- **Design of experiments to determine the hyperparameters of the SVM model**

To form a final learning machine for assigning servers to a new user of a set of users, we cannot use the 20 tuned SVMs from the previous section. Therefore, we must find a unique setting of the hyperparameters for all users. For this purpose, a design of experiments (DOE) is applied by which the minimum prediction accuracy among all 20 users' servers is maximized.

A total of 16 experiments have been conducted utilizing the design of experiments approach to determine the optimal hyperparameter values for the SVM model. In this regard, different values for the kernel, gamma, and C have been determined; moreover, multiple experiments have been conducted, incorporating various settings for each of the mentioned parameters. The outcomes of these experiments are presented in Table 7.

The results indicate that under the normal situation and with the parameters tested, the selection of the poly kernel has led to a minimum accuracy of 88% among users. Furthermore, a comprehensive analysis of the experiment's results has been conducted using MINITAB statistical analysis package, and the findings are visually presented in Figure 8.

Table 7. Design of experiments for hyperparameter setting

| ID | Kernel | Gamma | C | Min-accuracy (%) | | |
|----|--------|-------|---|--------|---------|-------|
|    |        |       |   | Normal | Special | Mixed |
| 1  | RBF     | 0.0001 | 1  | 79 | 56 | 64 |
| 2  | RBF     | 0.0001 | 10 | 88 | 79 | 77 |
| 3  | RBF     | 0.1    | 1  | 79 | 79 | 71 |
| 4  | RBF     | 0.1    | 10 | 79 | 79 | 71 |
| 5  | POLY    | 0.0001 | 1  | 86 | 70 | 72 |
| 6  | POLY    | 0.0001 | 10 | 88 | 85 | 81 |
| 7  | POLY    | 0.1    | 1  | 88 | 85 | 82 |
| 8  | POLY    | 0.1    | 10 | 88 | 85 | 82 |
| 9  | SIGMOID | 0.0001 | 1  | 36 | 19 | 41 |
| 10 | SIGMOID | 0.0001 | 10 | 44 | 32 | 31 |
| 11 | SIGMOID | 0.1    | 1  | 18 | 19 | 18 |
| 12 | SIGMOID | 0.1    | 10 | 18 | 19 | 18 |
| 13 | LINEAR  | 0.0001 | 1  | 87 | 84 | 72 |
| 14 | LINEAR  | 0.0001 | 10 | 87 | 84 | 73 |
| 15 | LINEAR  | 0.1    | 1  | 87 | 84 | 72 |
| 16 | LINEAR  | 0.1    | 10 | 87 | 84 | 73 |

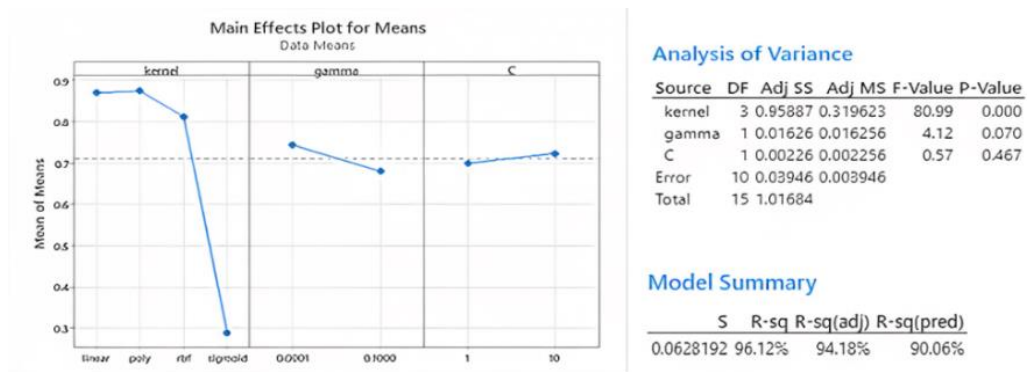

Fig 8. Analysis of the experimental design results for the normal situation

Figure 8 clarifies that the experiment's kernel and gamma factors are effective, but parameter C has little effect on the outcomes. The obtained results in this model are most affected by the kernel factor and least affected by the C factor. In the special situation, using the poly kernel leads to a minimum accuracy of 85%. On the other hand, implementing the linear kernel yields an accuracy of 84% (see Figure 9).

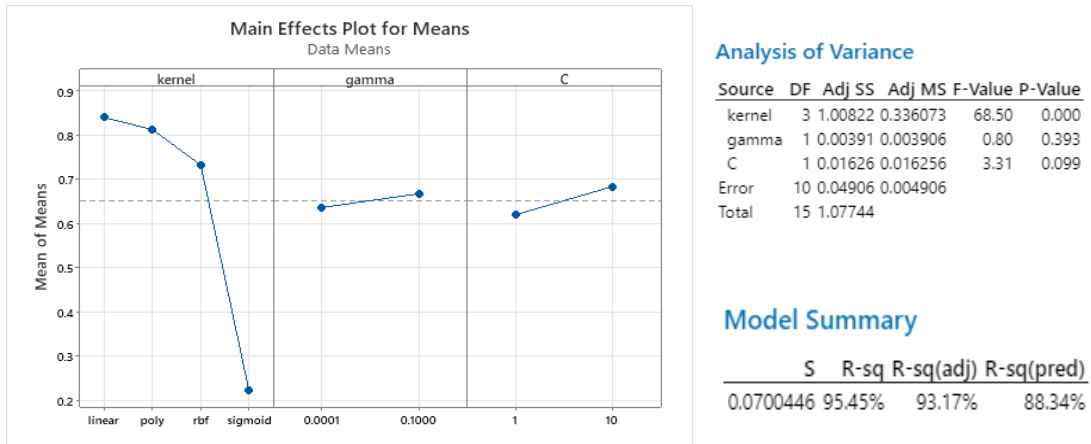

Fig 9. Analysis of the experimental design results for the special situation

As shown in Figure 9, kernel, C, and gamma are the effective factors, respectively. In this model, the kernel factor has the most impact, and the gamma factor has the least impact on the obtained results. Figure 10 shows that in the specific case, we have reached at least 82% accuracy among users by choosing the poly kernel.

Figure 10 clearly illustrates the impact of kernel, gamma, and C as effective factors in the model. Among these factors, the kernel factor has the most significant influence, while the C factor has the least impact on the obtained results. Using the Taguchi analysis method, the impact of these factors has been analyzed, and the C factor remains almost optimal at a level of 10 across all three modes. However, it does not significantly influence the overall results.

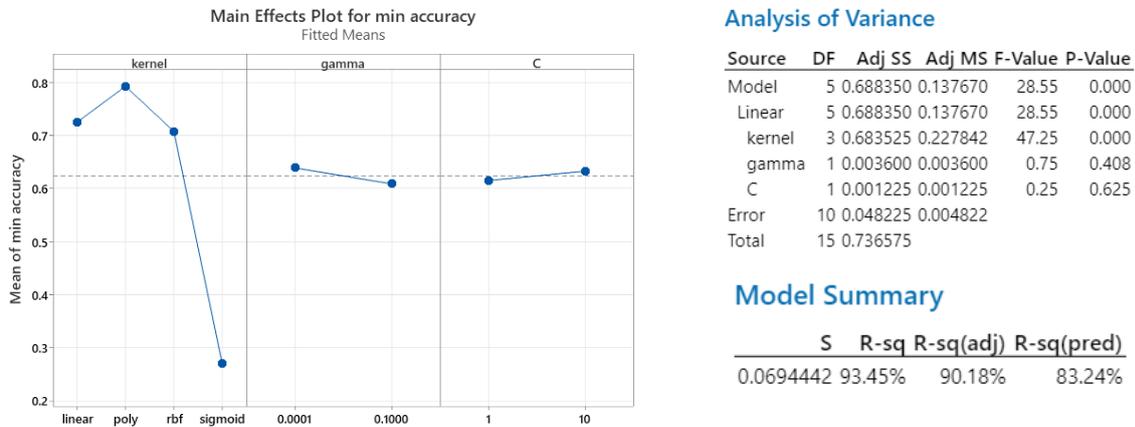

Fig 10. Analysis of the experimental design results for the mixed situation

The gamma factor is optimal at a level of 0.0001 in the normal situation, while for the special and mixed cases is not significant. So, we choose 0.0001 as our final setting for all the situations. The most influential factor remains the kernel factor.

The kernel factor is optimal at the poly level for the mixed situation. In the normal situation, the poly kernel performs slightly better than the linear kernel, but the linear kernel is optimal in the special case where other factors have less influence. But post-hoc analysis denotes that the main difference is due to the sigmoid setting, and the other three options did not significantly differ. So, we can keep the Poly kernel for all situations. Accordingly, the solution results for all users can be found in Table 8.

Table 8. The final setting of hyperparameters of all users' predictions using DOE

| Kernel | POLY | POLY | POLY |
|---|---|---|---|
| Gamma | 0.0001 | 0.0001 | 0.0001 |
| C | 10 | 10 | 10 |
| Accuracy (%) | Mixed | Normal | Special |
| User1 | 83 | 90 | 86 |
| User2 | 83 | 91 | 87 |
| User3 | 83 | 90 | 85 |
| User4 | 84 | 91 | 87 |
| User5 | 81 | 90 | 86 |
| User6 | 83 | 91 | 85 |
| User7 | 82 | 89 | 87 |
| User8 | 81 | 89 | 86 |
| User9 | 83 | 91 | 87 |
| User10 | 81 | 89 | 86 |
| User11 | 83 | 90 | 86 |
| User12 | 82 | 90 | 87 |
| User13 | 83 | 89 | 86 |
| User14 | 84 | 90 | 86 |
| User15 | 82 | 89 | 87 |
| User16 | 83 | 92 | 87 |
| User17 | 82 | 90 | 85 |
| User18 | 83 | 89 | 87 |
| User19 | 81 | 89 | 85 |
| User20 | 82 | 88 | 85 |
| **Min accuracy (%)** | **81** | **88** | **85** |
| **Average accuracy (%)** | **82.4** | **89.8** | **86.2** |

The problem has been successfully resolved using the SVM model, incorporating optimal fixed hyperparameters determined by analyzing the experimental design results. On average, the normal, special, and mixed modes have demonstrated superior performance, respectively.

### 4.3. Comparing the results of machine-learning models for the problem of placing applications in MEC servers

The performance of the models has been compared in terms of minimum accuracy and average accuracy, obtained by all three models (SVM, MLP in the case where grid search is used for parameter setting for each user, and SVM in the case where DOE finds the optimal hyperparameters). Their results are given in Table 9.

According to the results presented in Table 9, it is evident that the SVM-parameter model outperforms other models in both normal and special cases. In terms of the mixed mode, employing the MLP model to search for optimal parameters for each hyper user proves to be a superior approach. However, considering the marginal difference in results, utilizing the same SVM model with fixed parameters is feasible. Overall, the machine-learning model employed for application placement in edge servers achieves an impressive accuracy rate of over 80%. Among the myriad challenges encountered in the field of MEC, one of the most crucial ones revolves around reducing user connection delays to servers. Furthermore, when addressing optimization problems in high-dimensional spaces and diverse uncertain scenarios, a notable obstacle arises in costly calculations and extensive time requirements. The time

CPLEX takes to solve the MEC model across various dimensions is comprehensively outlined in Table 10.

Table 9. Comparison of performance of models

| Model | Situation | Min accuracy (%) | Average accuracy (%) |
|---|---|---|---|
| SVM | Normal | 85 | **90** |
| MLP | Normal | 76 | 89 |
| SVM-DOE | Normal | **88** | 90 |
| SVM | Special | 80 | 84 |
| MLP | Special | 81 | 83 |
| SVM-DOE | Special | **85** | **86** |
| SVM | Mixed | 79 | 83 |
| MLP | Mixed | **82** | **85** |
| SVM-DOE | Mixed | 81 | 82 |

Table 10. Time to solve the application placement model in edge servers by CPLEX in different dimensions

| Model | Number of users | Number of servers | Number of scenarios | Solution time (s) |
|---|---|---|---|---|
| 1 | 10 | 3 | 15 | 0.329 |
| 2 | 20 | 5 | 25 | 391 |
| 3 | 30 | 7 | 35 | 2922 |
| 4 | 40 | 10 | 45 | >10000 |
| 5 | 50 | 13 | 55 | >10000 |
| 6 | 60 | 15 | 70 | >10000 |

It is evident that solving precise mathematical problems in high dimensions requires a significant amount of time. However, the time limitations become problematic when considering a mode where the application needs to be executed every few minutes by twenty users across five servers over eight hours (equivalent to 480 minutes). Particularly with the growing number of scenarios in the second stage, the solution time required becomes very high. For instance, the first case of Table 11 has been examined. If the model needs to be updated every five minutes, it demands 626 minutes to run, exceeding the available 480 minutes. As the number of scenarios increases, the time needed to solve the model also escalates. In contrast, the time for solving and updating the trained models remains under one second during these mentioned intervals.

Table 11. Comparison of CPU time (seconds) for optimization runs in 8 working hours performed by GAMS and

| Model | Number of users | Number of servers | Number of scenarios | Number of runs required | | | | | Method |
| | | | | 1 (Every 480 minutes) | 32 (Every 15 minutes) | 48 (Every 10 minutes) | 60 (Every 8 minutes) | 96 (Every 5 minutes) | |
|---|---|---|---|---|---|---|---|---|---|
| 1 | 20 | 5 | 25 | 391 (Sec) | ۱۲۵۱۲=۳۲*۳۹۱ | ۱۸۷۶۸ | ۲۳۴۶۰ | ۳۷۵۳۶ | GAMS (Cplex) |
| 2 | 20 | 5 | 50 | 718.5 | 22992 | 34488 | 43110 | 68976 | |
| 3 | 20 | 5 | 80 | 3313.7 | 106038.4 | 159057.6 | 198822 | 318115.2 | |
| 4 | 20 | 5 | 80 | 0.00149 | 0.0475 | 0.093 | 0.126 | 0.156 | ML |

In the real world, making even remotely close to optimal decisions can lead to substantial improvements in service quality and cost reduction. Notably, the time required to solve a machine-learning model, both during training and actual usage, is incredibly short and yields answers close to the optimal solution. To illustrate this, the time spent to train and test the SVM-DOE model on a dataset of 1800 records can be noted. In three distinct situations, namely special, normal, and mixed, the entire process took approximately 30-35 seconds. During this short time, the learning model achieved an impressive performance level of over 80%.

Furthermore, once the model is trained, the time required for its utilization can be reduced to less than a second, as simple as a click. Consequently, while machine-learning models may not provide the optimal answer, they offer solutions that are remarkably close to it, all while saving considerable time and money. Given that users of mobile devices in the MEC network may relocate after the initial placement of an application, the implementation of MLP and SVM algorithms becomes crucial. These algorithms enable us to understand the relationship between various input parameters, such as the distances of individual users from the servers, the volume of their requests, and the optimal values of decision variables.

## 5. Conclusion and future works

Edge computing has emerged as a complementary solution to address the limitations of cloud computing. It comprises a network of compact data centers that manage and process data locally. Edge computing alleviates congestion and reduces network traffic by minimizing the need to transmit all data to a central server (Asheralieva & Niyato, 2019). However, optimizing Mobile Edge Computing (MEC) poses challenges, as traditional programs such as GAMS and CPLEX employ time-consuming algorithms to solve complex problems with high dimensions and uncertain scenarios. Leveraging machine-learning algorithms and problem-solving records can significantly reduce solving time to overcome these hurdles and achieve optimal server allocation. This study investigates the potential of employing machine-learning models for operational decision-making, specifically short-term decisions. Two machine-learning models have been used within a stochastic programming approach. These models have shown the ability to learn from a small set of optimal solutions provided as a training dataset.

Moreover, the results obtained from the study have highlighted the SVM model's proficiency in making efficient decisions regarding allocating user-requested applications to available servers. Industries can readily utilize these models as software by simply incorporating server and user location data, calculating distances between them, and assessing the daily volume of user requests. It is important to note that while machine-learning solutions may not be as optimal as those derived from exact solvers, they offer near-optimal solutions in significantly less time. Consequently, all machine-learning models can be promptly evaluated with new samples. Accordingly, future research endeavors could explore the utilization of clustering techniques on user location datasets from various days to ascertain whether the system is in a normal or specialized mode. Based on this determination, the model can select the most appropriate approach. Additionally, employing real datasets with larger dimensions could enhance the analysis, while machine-learning models have the potential to predict the ideal value of the objective function. Moreover, evaluating the model's accuracy, fine-tuning hyperparameters, and exploring more sophisticated prediction models are all worth exploring.